# Task-Aware Tuning of Time Constants in Spiking Neural Networks for Multimodal Classification

Chiu-Chang Cheng[1], Kapil Bhardwaj[2], Ya-Ning Chang[1], Sayani Majumdar[1,2,*] and Chao-Hung Wang[1,3,*]

[1]Miin Wu School of Computing, National Cheng Kung University, 70101 TainanNo City, Taiwan (R.O.C.)

[2]Faculty of Information Technology and Communication Sciences, Tampere University, 33720 Tampere, Finland

[3]Academy of Innovative Semiconductor and Sustainable Manufacturing, National Cheng Kung University, 70101 Tainan City, Taiwan (R.O.C.)

Corresponding authors: sayani.majumdar@tuni.fi; chwang@gs.ncku.edu.tw

**Abstract**

Spiking Neural Networks (SNNs) are promising candidates for low-power edge computing in domains such as wearable sensing and time-series analysis. A key neuronal parameter, the leaky time constant (LTC), governs temporal integration of information in Leaky Integrate-and-Fire (LIF) neurons, yet its impact on feedforward SNN performance across different data modalities remains underexplored. This study investigates the role of LTC in a temporally adaptive feedforward SNN applied to static image, dynamic image, and biosignal time-series classification. Presented experiments demonstrate that LTCs critically affect inference accuracy, synaptic weight distributions, and firing dynamics. For static and dynamic images, intermediate LTCs yield higher accuracy and compact, centered weight histograms, reflecting stable feature encoding. In time-series tasks, optimal LTCs enhance temporal feature retention and result in broader weight sparsity, allowing for tolerance of LTC variations. The provided results show that inference accuracy peaks at specific LTC ranges, with significant degradation beyond this optimal band due to over-integration or excessive forgetting. Firing rate analysis reveals a strong interplay between LTC, network depth, and energy efficiency, underscoring the importance of balanced spiking activity. These findings reveal that task-specific LTC tuning is essential for efficient spike coding and robust learning. The results provide practical guidelines for hardware-aware SNN optimization and highlight how neuronal time constants can be designed to match task dynamics. This work contributes toward scalable, ultra-low-power SNN deployment for real-time classification tasks in neuromorphic computing.

## Introduction

Modern intelligent systems increasingly rely on real-time sensing and on-device decision-making in applications like health monitoring and autonomous navigation. These edge applications demand low-latency, energy-efficient computing, which traditional cloud-based models struggle to support due to bandwidth and power constraints [1]. Spiking Neural Networks (SNNs) offer a biologically inspired alternative, featuring event-driven processing, sparse activity, and inherent temporal coding [2]. Neuromorphic platforms like Intel’s Loihi have demonstrated SNNs' energy advantages over conventional ANNs [3], particularly in edge health monitoring tasks using learning rules like SLAYER [4]. Adaptive SNN architectures further enhance temporal modeling for time-series data [5], positioning SNNs as a key enabler for energy-efficient edge intelligence [6].

SNN-based edge computing has emerged as a solution to process data locally, minimizing communication energy and latency overhead and improving energy efficiency. Near-sensor and in-sensor computing modalities aim to shift computation closer to data acquisition points, significantly reducing redundant data transfer and latency [7].

However, for SNNs to perform optimally on temporal tasks such as electrocardiogram (ECG) data classification, careful tuning of their internal parameters is essential [8]. One such critical parameter is the leaky time constant (LTC), which governs how long a neuron integrates incoming spikes before decaying. A small LTC causes fast forgetting of past inputs, while a large LTC retains older information; both extremes can hinder the network's performance depending on the application. Unfortunately, most SNN studies have either used fixed LTC values or manually tuned them without systematic optimization, creating challenges in both performance consistency and hardware implementation [9, 10].

On the hardware side, numerous electronic neurons and synapses have demonstrated the capability to tune LTCs. For instance, to the best of our knowledge, silicon neurons [11], electrochemical iontronic synapses (EIS) [12], single logic transistors [13], $NbO_x$ memristors [14], diffusive memristors [15], ferroelectric tunnel junction (FTJ) and ferroelectric field effect transistor (FeFET) memristors [16, 17], ferroelectric tunnel memristors [18], two-dimensional materials [19], optoelectronic memristors [20], and triboelectric or charge-trapping transistors [21, 22] have all demonstrated strong tunability of LTCs, although each device has its own optimal operating and tuning window. This diversity offers promising pathways for task-specific circuit optimization, but it also necessitates clearer guidelines linking algorithmic requirements to hardware design.

Though solid-state device-based neurons and synapses have gained significant popularity in recent years [11-13, 16-20, 22], the use of CMOS technology remains indispensable due to its maturity, scalability, and compatibility with standard fabrication processes. CMOS Leaky Integrate-and-Fire (LIF) neuron circuits can emulate a wide range of leaky time constants ($\tau$) by employing several well-established circuit techniques [11, 23-28]. One of the most common methods is to use a Metal-Oxide-Semiconductor Field-Effect-Transistor (MOSFET)-based current sink to implement a variable leak path, enabling exponential decay of the membrane potential, where the rate of decay is controlled by adjusting the gate bias or current source parameters [11, 23]. Additionally, the membrane capacitance ($C_{mem}$) can be varied using digitally programmable capacitor arrays or switched-capacitor techniques, allowing precise tuning of $\tau$ based on the relationship $\tau=C_{mem}/g_{leak}$ (with $g_{leak}$ being the tunable conductance via tunable subthreshold MOSFET) [24]. Operating transistors in the subthreshold region further enables biologically realistic, long-duration time constants due to the exponential current-voltage characteristics of MOS devices, all while consuming very low power [23, 25]. In advanced neuromorphic systems such as Loihi and DYNAPs, digital-to-analog converters (DACs) are used to program leak and integration currents in real time, offering adaptive and configurable $\tau$ for diverse computational needs [26, 27]. To ensure reliable performance across process, voltage, and temperature variations, current mirrors and temperature-compensated bandgap reference circuits are also employed for bias stabilization [28]. These combined techniques make CMOS LIF neurons highly tunable, robust, and efficient for large-scale neuromorphic computing platforms.

Some studies on the software side have investigated the impact of leaky time constants on SNN performance. For example, the BA-LIF (Brain-Inspired Leaky Integrate-and-Fire) neuron model presented in [29] achieves high accuracy without relying on time-consuming, manual initialization of the membrane time constant. This reflects a broader trend in SNN research, moving away from static parameter settings toward more adaptive, learnable configurations to better suit diverse temporal tasks. More recent studies emphasized on the importance of moving beyond fixed membrane time constants ($\tau_m$), as static configurations often limit a network's ability to model variable temporal dependencies. Zeng et al. [30] demonstrated that using small, fixed $\tau_m$ values can reduce temporal lag and enhance both convergence speed and classification accuracy in recurrent SNNs, particularly for tasks with short-term dynamics. However, such fixed strategies may not generalize well across tasks with diverse temporal characteristics. To address this, Pazderka et al. [31] introduced per-neuron learnable $\tau_m$, enabling individualized temporal integration profiles that align more closely with data-specific structures, thereby improving training stability and task performance. Building on this idea, Zhang et al. [32] proposed the Dual Adaptive Leaky Integrate-and-Fire (DA-LIF) model, which decouples spatial and temporal decay with independent adaptive parameters, allowing more flexible spatiotemporal encoding while reducing inference latency. Li et al. [33] examined synaptic time constants ($\tau_s$), showing that faster synaptic decay deepens temporal abstraction and influences memory retention in spiking circuits. Moro et al. [34] extended the concept by employing hierarchical $\tau_m$ settings across layers to facilitate multiscale temporal feature extraction. Other models such as Adaptive Spiking Recurrent Neural Networks (ASRNNs) [35] and learnable-$\tau_m$ strategies by Fang et al. [36] further highlight the advantages of dynamic temporal tuning in complex time-series tasks. Complementary approaches like Datta et al.'s [37] dynamic time stepping in vision transformers, and hardware-aware techniques by Bhaskarabhatla et al. [38] and Rathi and Roy [39], underscore the need for coordinated algorithm-hardware optimization of leaky time constants. Together, these works reinforce that LTC adaptation is a crucial dimension in optimizing SNNs for efficient and scalable performance; however, a unified framework for hardware-software co-design remains lacking. Specifically, a Design-Technology Co-Optimization (DTCO) approach that connects algorithmic LTC tuning with hardware constraints is needed. Understanding how LTC impacts different task modalities, static images, dynamic vision, and time-series signals, can guide both circuit design and model training. This study aims to fill that gap by systematically evaluating the role of LTCs across three domains. The results show that inference accuracy peaks within specific LTC ranges and degrades significantly outside this optimal band due to over-integration or excessive forgetting, providing a practical guideline for LTC-aware neuromorphic system design.

## Experimental details

In this work, a fully connected spiking neural network (SNN) is constructed using LIF neurons with adjustable LTCs. The Direct-Input-Encode method is employed to convert input values into spike trains [39]. Three classification tasks are used to evaluate the impact of LTCs: (1) static image classification, (2) dynamic image classification, and (3) time-series data classification. For static image classification, handwritten digits from the MNIST dataset are directly fed into a 2-layer SNN with 10 output classes. The LIF pre-neurons generate spikes once their membrane potential exceeds the threshold, propagating signals to post-neurons. For dynamic image classification, MNIST images are partitioned into four sequential segments and fed into the SNN across time steps, enabling analysis of temporal information processing. For time-series data classification, the SNN architecture is modified to include a single input neuron and four network layers with four output classes, reflecting the higher temporal complexity of time-series tasks. Similar to the previous cases, input signals are directly fed into the LIF-based SNN, where spike trains are inherently generated and propagated through the network. Detailed SNN configurations are provided in the **Supplementary Materials**.

## Results and Discussion

Neuronal time constants represent the leakage behavior of the neuron's membrane potential, which is believed to encode the history of spike information. In software, neuronal time constants can be implemented by directly assigning different time constant values to LIF neurons. In hardware, numerous electronic neuron devices have demonstrated specific time constants and the tunability of these constants. The correlation between the software and hardware approaches is presented, and the implementation of neuronal time constants is illustrated in Figure 1.

On the hardware side, neurons can be implemented using various electronic devices, such as CMOS and non-CMOS transistors, memristors, and optoelectronic memristors (Table 1). To the best of our knowledge, Table 1 lists representative time constant values collected from the literature. These values span a wide range, from $10^{-3}$ to $10^{2}$ seconds, regardless of the type of device, transistor, or memristor.

On the software side, a time-constant-aware Spiking Neural Network (τ-SNN) model is established to assess the effects of the wide variability in electronic neuron device time constants through a software-based approach. The conversion of time constants between the software and hardware approaches is carried out using the following equation, Equation 1:

$$\tau_{software} = \frac{\tau_{hardware}}{\Delta t} \tag{1}$$

Here, $\tau_{software}$ represents the time constant in software simulations, while $\tau_{hardware}$ denotes the time constant in hardware devices. Taking ECG data as an example, the sampling rate of ECG is 360 Hz, and the software time constant $\tau_{software}$ is set to 64. The corresponding hardware time constant is 64 divided by 360 seconds, which equals approximately 0.1777 seconds. Table 2 lists the conversions of time constants between software and hardware domains, with values ranging from $10^{-3}$ to 1 second, aligning with those of practical electronic devices. Detailed conversion procedures are provided in the **Supplementary Materials.**

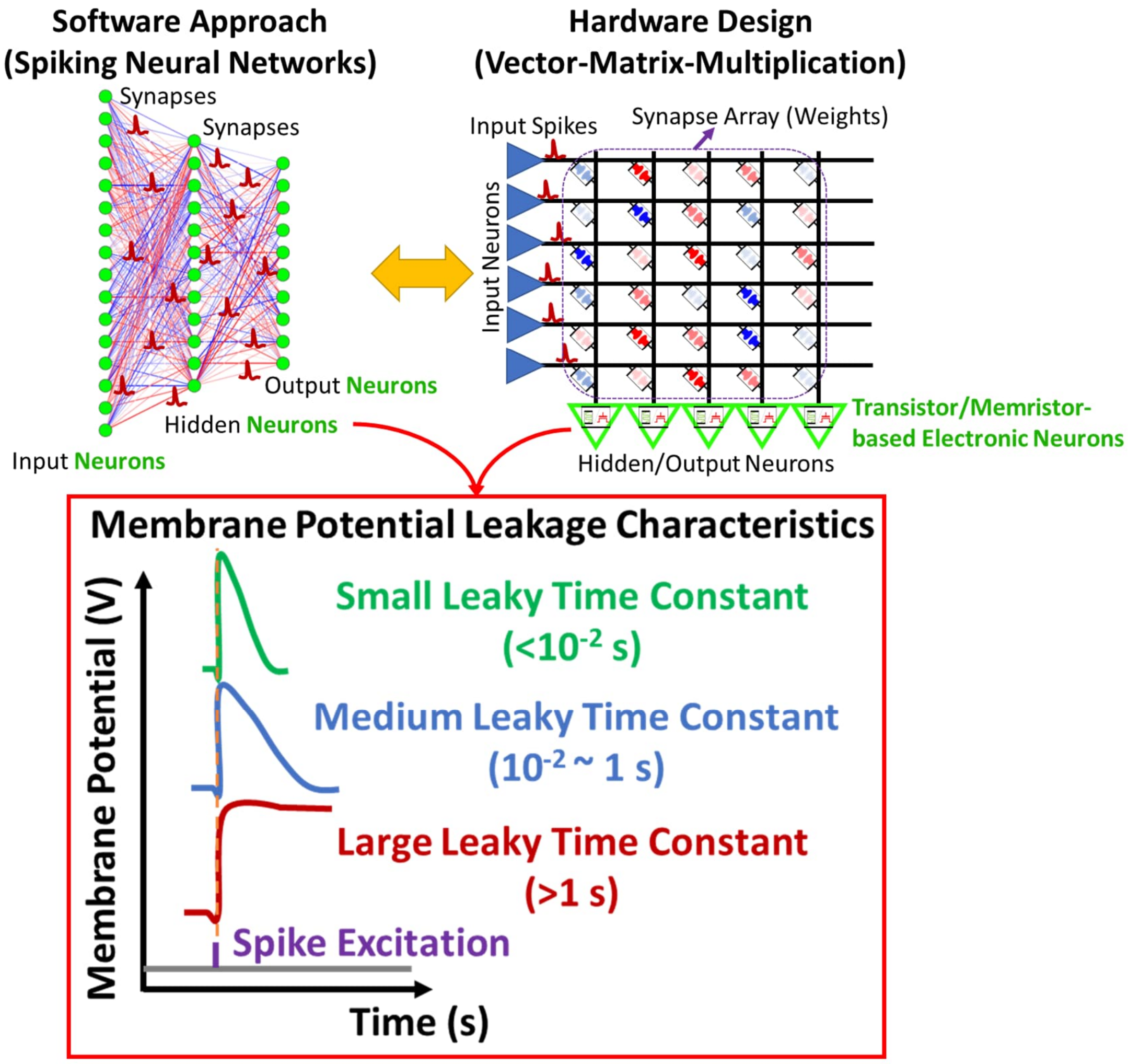

Figure 1. The conceptual framework of hardware-software co-optimization, or Design-Technology Co-Optimization (DTCO), accounts for the impact of neuron membrane potential leakage characteristics on both software and hardware implementations

**Table 1.** Comparison of electronic neuronal devices

| Device | Materials | Time Constant (s) | Decay Mechanism/Model | Operation Techniques | Plasticity Type | Learning Rule Support | Reference |
|---|---|---|---|---|---|---|---|

| | | | | | | | | |
|---|---|---|---|---|---|---|---|---|
| **Transistor-based Neuronal Devices** | **High-k $HfO_2$ Transistor** | TiN/$HfO_2$/Si channel | 2.93 $\times 10^{-3}$ | Shallow and deep hole traps | Voltage amplitude, pulse width/freq | Short-Term Potentiation (STP) + Long-Term Potentiation (LTP) (by voltage level) | Spike-Time Dependent Plasticity (STDP), Spike-Rate Dependent Plasticity (SRDP) | [13] |
| | **Ferroelectric FET** | Au/P(VDF-TrFE)/Pentacene | 1.66 – 6.97 | Tribo-charge depletion | Pulse intermittence/frequency | STP | STDP (limited) | [16] |
| | **Triboelectric Charge-trap Transistor** | $MoTe_2$ + $Al_2O_3$/$HfO_2$ | 1.57 – 83.94 | Charge trapping in oxide stack | Mechanical + electrical pulses | LTP (nonvolatile) | Hebbian, Conditioning | [22] |
| | **$MoS_2$ Neural** | Ionic liquid/$MoS_2$ | 29.25 | Mixed electronic–ionic decay | Gate pulse amplitude/frequency | STP + LTP (via amplitude) | STDP, Paired-Pulse Facilitatio | [19] |
| | **Device** | | | | | | n (PPF), Hebbian | |
| | **Li-based Synaptic Transistor** | LiPON/$LiCoO_2$ | 166.06 | Li-ion intercalation | Gate voltage, ion conc. | STP + LTP (pulse width) | STDP-like, PPF | [12] |
| | **Li-ion Solid Electrolyte 2D α-$MoO_3$ Nanosheet Transistor** | $LiClO_4$/α-$MoO_3$ | 19.95 | $Li^+$ doping/undoping | Gate pulse amplitude | STP + LTP | Hebbian, STDP | [12] |
| | **$MoS_2$/$Na^+$-diffused** | $Na^+$:$SiO_2$/$MoS_2$ | 21.99–69.26 | Na-ion diffusion | Gate pulse width/freq | STP + LTP (at high T) | STDP, Long-Term Potentiation(LTP)/ | [12] |
| | **$SiO_2$ Transistor** | | | | | | Long-Term Depression(LTD) | |
| | **Na-based $WO_x$ Synaptic Transistor** | $NaWO_3$/$WO_x$ | 221.77 | Na-ion drift in oxide | Voltage pulse duration/freq | STP + LTP | STDP, PPF | [12] |
| | **Standard CMOS transistors** | Standard CMOS LIF Neuron (analog, current-mode) | $10^{-3}$ – 1s (tunable) | Exponential decay via RC membrane (capacitor + programmable current sink or MOS resistor) | Varying membrane capacitance and leak conductance (bias-controlled), tunable current mirror sinks | STP (integration); external circuits for LTP/STDP | STDP-compatible (via synaptic circuits), R-STDP, Hebbian | [11] |
| **Memristor-based Neuronal Devices** | **$TiO_2$:ZnO QD Me** | Ag/$TiO_2$:ZnO/FTO | 0.15 – 0.36 | Photocarrier recombination | Light + voltage pulses | STP (photo) + LTP (electrical) | Pavlovian conditioning, Hebbian | [20] |
| | **mristor** | | | | | | | |
| | **Ferroelectric Memristor** | Pt/$BaTiO_3$/Nb:STO | 8.69 $\times 10^{-2}$ | Polarization + trap migration | Voltage pulse + freq | STP + LTP | STDP, Hebbian | [17] |
| | **Organic Ferroelectric FTJ Memristor** | Au/P(VDF–TrFE)/Nb:STO | 39–118 | FE depolarization relaxation | Pulse timing/freq | STP + LTP | STDP, Hebbian | [18] |

Table 2. Conversion of time constants from the discrete domain to the continuous domain

| Time constants in the software, discrete domain | Time constants in the hardware, continuous domain (sec) |
|---|---|
| 2 | 0.0056 |
| 4 | 0.0111 |
| 8 | 0.0222 |
| 16 | 0.0444 |
| 32 | 0.0889 |
| 64 | 0.1778 |
| 128 | 0.3556 |
| 256 | 0.7111 |
| 512 | 1.4222 |

To evaluate the impact of neuronal time constants on the performance of τ-SNN, network architectures are carefully designed for different tasks. Detailed descriptions of these architectures are provided in the **Supplementary Materials**. Three major tasks are evaluated: static image classification, dynamic image classification, and time-series data classification, as shown in Figure 2. Static image classification establishes the baseline performance of τ-SNN, while dynamic image classification highlights its strengths in processing time-dependent information. Time-series data classification further investigates the complexities of real-world tasks involving inherent temporal correlations. In τ-SNN, memory functionality may arise not only from synapses but also from neurons through the LIF mechanism, underscoring the importance of identifying an optimal time constant.

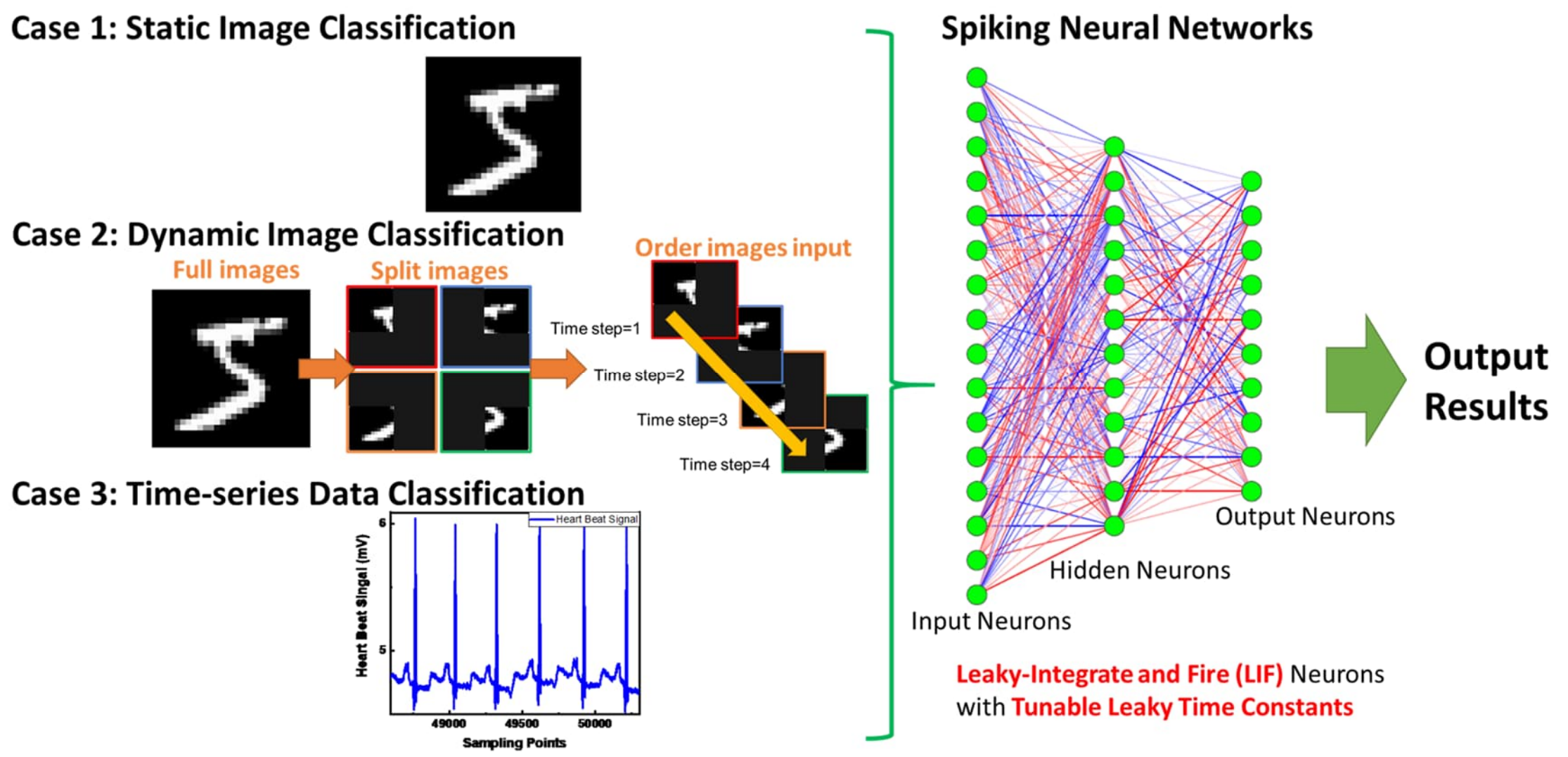

Figure 2. The workflow for investigating the impact of time constants on Spiking Neural Network performance across three scenarios using a software-based approach

Figure 3 demonstrates the impact of neuronal time constants on the performance of τ-SNN in static and dynamic image classification tasks. Under various time constant settings, τ-SNN can be successfully trained for both tasks. However, when the neuronal time constants used during inference do not match those used during training, discrepancies in network performance emerge in both tasks.

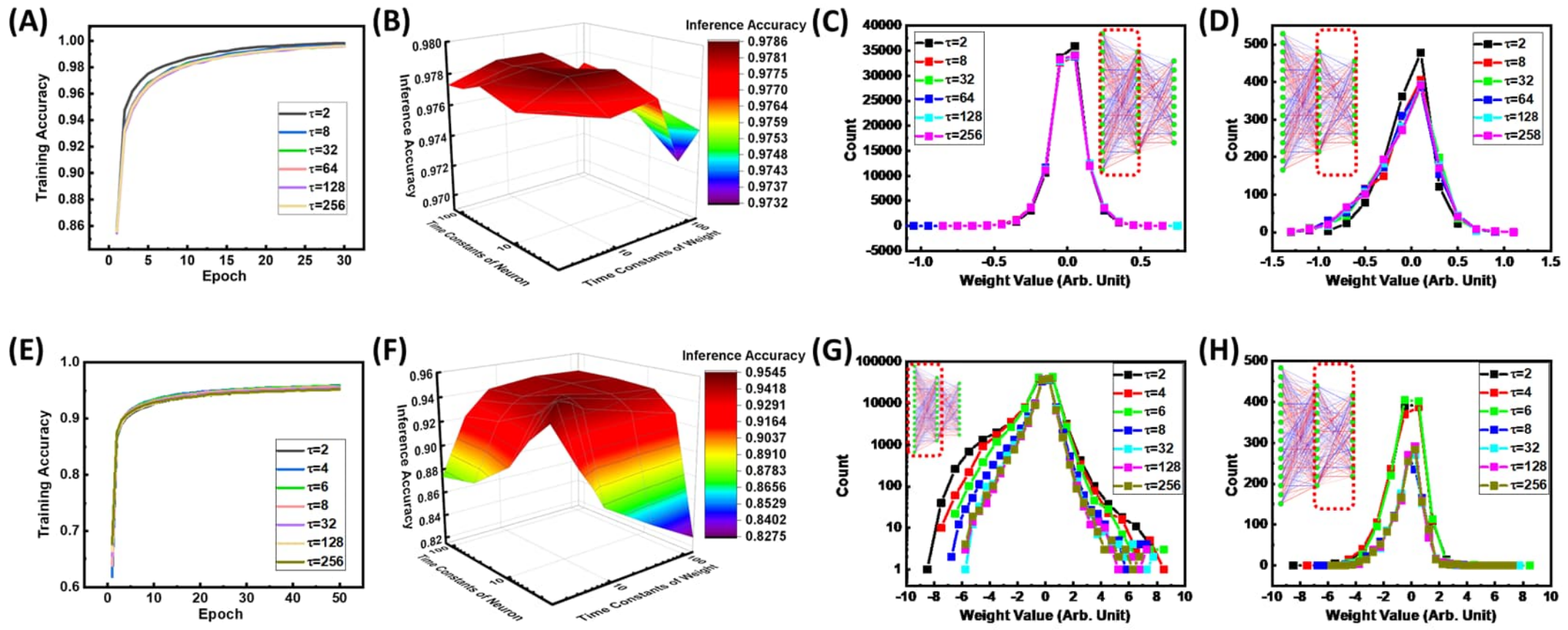

Figure 3. Impact of τ-SNN performance on static and dynamic image classification. (A–D) Training history, inference accuracy, first-layer weights, and second-layer weights for static image classification. (E–H) Training history, inference accuracy, first-layer weights, and second-layer weights for dynamic image classification.

In the static image classification task, network performance does not significantly degrade (with inference accuracy remaining above 97%) when there is a mismatch between the inference and training time constants. In contrast, the dynamic image classification task is more sensitive to such mismatches, with accuracy dropping noticeably to 82% during inference. To further investigate this result, the weight distributions of τ-SNN under different time constant conditions are analyzed, as shown in Figure 3C–3D and Figure 3G–3H. In the static image classification task, the weight distributions in the first and second layers of τ-SNN remain relatively unaffected by different time constants. However, in the dynamic image classification task, the weight distribution becomes more concentrated as the time constant increases. This can be explained by the fact that larger time constants allow neurons to retain spike history for longer periods. As a result, the weight distribution shifts toward zero, indicating that information storage shifts from synapses to neurons. Additionally, since temporal information is encoded in the neurons of τ-SNN, mismatches in time constants have a more pronounced effect on accuracy in dynamic tasks.

In addition, to further examine the ability of τ-SNN to process time-dependent correlations, its performance on ECG classification has been evaluated. Figure 4 illustrates the inference accuracy on heartbeat data, the weight distributions across each neural network layer, and the firing rates in each layer. The inference accuracy exhibits a trend similar to that observed in the dynamic image classification task, but it is more significantly affected by the mismatch between inference and training time constants, resulting in a narrower window for time constant tuning (Figure 4A).

The weight distributions from the first to the last layer of τ-SNN (Figure 4B–E) show that only the first and last layers exhibit broader distributions as the time constant increases, while the intermediate layers maintain similar distributions. This trend contrasts with that observed in the dynamic image classification tasks. It is suggested that the ECG classification task poses greater challenges for τ-SNN in terms of learning and achieving high accuracy within an acceptable time constant range. As shown in Figure 4A, the time constant window for achieving high accuracy is broader when using larger time constants, but becomes extremely narrow with smaller time constants. This indicates that smaller time constants lead to faster information decay. Although high accuracy can still be achieved with an optimal weight distribution, the tolerance to time constant variation decreases due to the tighter weight distribution. In this case, while the neuron is responsible for memory, it becomes more sensitive to variation, whereas the synaptic weights remain relatively robust. Conversely, for larger time constants, the trend reverses. The impact of time constant variation is reduced due to the broader weight distribution. As a result, although neurons contribute to memory retention through increased time constants, they require broader weight distributions to mitigate the effects of variation, ultimately relying on the stability of synaptic weights.

The firing rates across each layer and time constant, shown in Figure 4F, illustrate how spike information propagates through τ-SNN. In general, the firing rate increases with layer depth, indicating that spike information is progressively processed and amplified within the network. Regarding time constant dependency, a larger time constant reduces the likelihood of spike emission in response to presynaptic input, resulting in lower firing rates, an effect confirmed by our results.

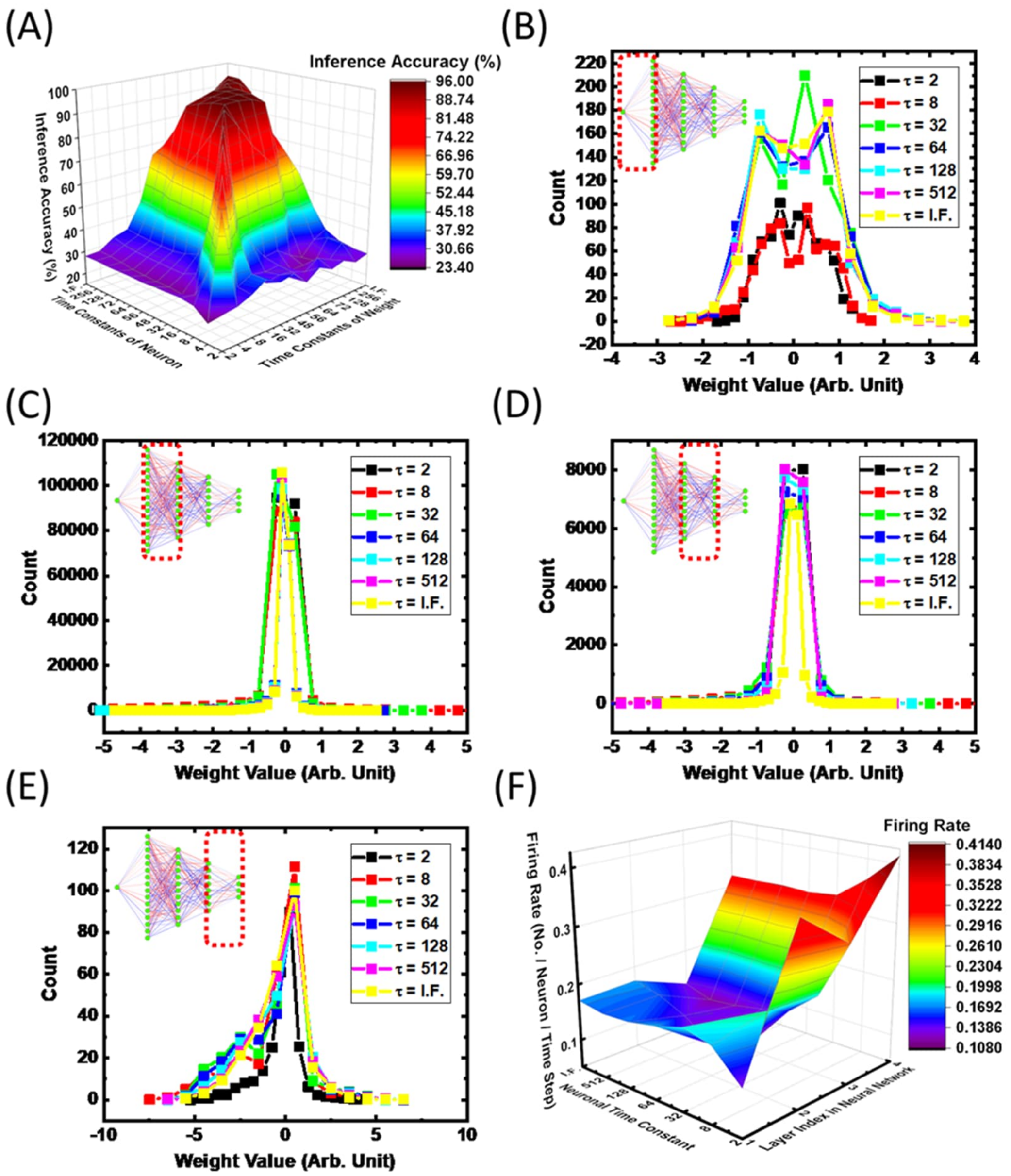

Figure 4. (A) Inference accuracy. (B–E) First-layer, second-layer, third-layer, and fourth-layer weights for time-series data classification. (F) Firing rates across different neuronal time constants and network layers in the τ-SNN.

Based on our results, a clear guideline can be established for designing an optimal electronic neuron device when designing an SNN chip. Figure 5 depicts neuronal time constants reported from CMOS-based or emerging non-volatile or volatile memory-based electronic devices. For an SNN chip targeting static image classification tasks, any of the listed devices can be used, as the τ value has limited impact on accuracy. However, for an SNN chip targeting dynamic image classification tasks, a relatively small time constant is preferred, i.e., τ > 4 or 0.01 seconds. Here, only the high-k $HfO_2$ transistor in Table 1 fails to meet the requirement. For an SNN chip targeting ECG classification tasks, considering the maximum accuracy window with respect to time constant variation, a large time constant is highly recommended, i.e., τ > 72 or 0.2 seconds. In this case, most of the devices listed in Table 1 meet this criterion without any changes in the device geometry and parameters. For the high-k $HfO_2$ transistor and the ferroelectric memristor reported in literature [13, 17, 18], device designs or stack geometry need modification to match the time constant requirement.

Considering the device physics, high-k $HfO_2$ transistors are logic switches with ultrafast operation. The rest of the transistor- or memristor-based devices have volatile or non-volatile memory properties that make them suitable as LIF neurons. Since the impact of LTC in an SNN has not been studied extensively, designing such parameters for a hardware neuron has not been considered very much. In the present scenario, the modulation of the LTC can be obtained in hardware by modulating the leaky part of the neuron device or circuit. For transistors, it can be achieved by utilizing charge trapping properties in the dielectric gate stack or dielectric-semiconductor interface. Traps in the gate dielectric can capture free charge carriers from the channel, effectively immobilizing them. The process of trapping and detrapping charge carriers is not instantaneous. It can happen over various timescales, from very fast of the order of microseconds to very slow of the order of seconds or even longer. This slow response to changes in gate voltage can lead to slower transistor operation and longer LTC [40]. By engineered creation of trap centres, it is possible to have larger LTCs in these devices. For Ferroelectric diodes, tunnel junction memristors, or FeFETs, the stability of the polarization charge determines the LTC. It is possible to modify screening charge carrier density of the electrode [41] or insert a thin dielectric layer and modifying the thickness of the ferroelectric or dielectric layer or dielectric constants of the materials to design different time constants of decay in the devices suitable for a specific application [42].

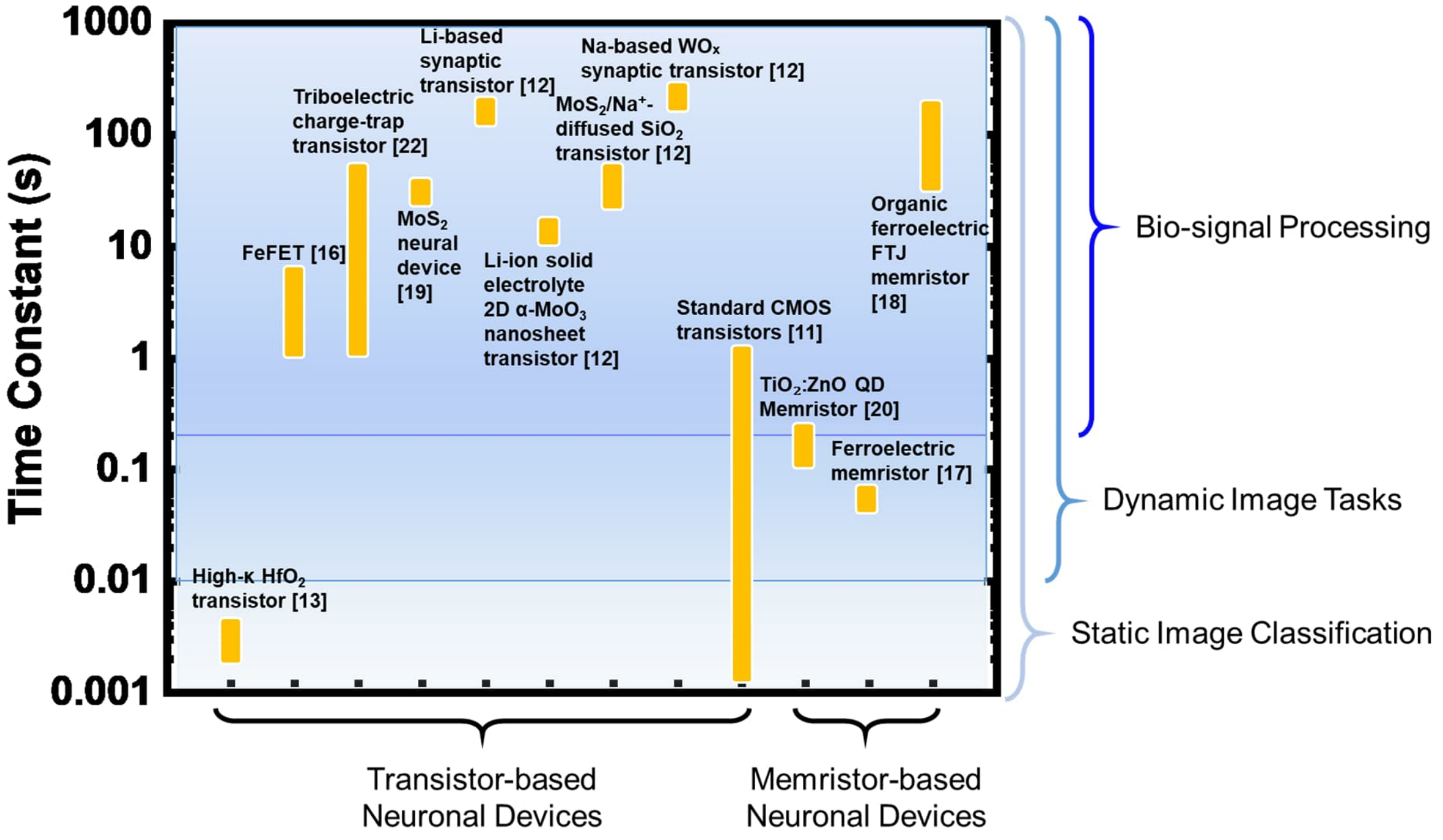

Figure 5. Time constant landscape of electronic neuronal devices across three application scenarios, illustrating optimal device behavior in different operational contexts.

## Conclusions

This study systematically investigated the influence of neuronal leaky time constants on the inference performance of τ-SNNs across three representative tasks: static image classification, dynamic image classification, and ECG-based time-series classification. Through a software-level evaluation framework, we demonstrated that leaky time constants (LTCs) critically affect the temporal integration behavior of neurons and must be carefully tuned based on task characteristics. Our findings suggest that for static image classification, τ-SNNs exhibit robustness to mismatches between training and inference time constants, indicating minimal sensitivity to LTC variations. In contrast, dynamic image classification tasks show significant performance degradation when small LTCs are used, as the network fails to retain sufficient temporal context. This degradation is reflected in the synaptic weight distribution, which becomes increasingly centered near zero, suggesting that neurons begin compensating by implicitly acting as memory units. For time-series tasks such as ECG classification, τ-SNNs are particularly sensitive to time constant mismatches, with larger LTCs ($\tau > 72$ or ~0.2 seconds) offering improved accuracy and robustness against variability. Based on these insights, we provide a task-specific guideline for selecting suitable electronic neuron devices with optimal

LTC tuning characteristics, offering valuable design considerations for future neuromorphic systems.

## References

[1] X. Xu, S. Zang, M. Bilal, X. Xu, and W. Dou, "Intelligent architecture and platforms for private edge cloud systems: A review," *Future Generation Computer Systems,* vol. 160, pp. 457-471, 2024/11/01/ 2024, doi: https://doi.org/10.1016/j.future.2024.06.024.

[2] E. Kim and Y. Kim, "Exploring the potential of spiking neural networks in biomedical applications: advantages, limitations, and future perspectives," (in eng), *Biomed Eng Lett,* vol. 14, no. 5, pp. 967-980, Sep 2024, doi: 10.1007/s13534-024-00403-1.

[3] M. Davies *et al.*, "Advancing Neuromorphic Computing With Loihi: A Survey of Results and Outlook," *Proceedings of the IEEE,* vol. 109, no. 5, pp. 911-934, 2021, doi: 10.1109/JPROC.2021.3067593.

[4] S. B. Shrestha and G. Orchard, "SLAYER: Spike Layer Error Reassignment in Time," 2018. [Online]. Available: https://proceedings.neurips.cc/paper_files/paper/2018/file/82f2b308c3b01637c607ce05f52a2fed-Paper.pdf.

[5] B. Yin, F. Corradi, and S. M. Bohté, "Accurate and efficient time-domain classification with adaptive spiking recurrent neural networks," *Nat. Mach. Intell.,* vol. 3, no. 10, pp. 905-913, / 2021, doi: 10.1038/S42256-021-00397-W.

[6] K. Roy, A. Jaiswal, and P. Panda, "Towards spike-based machine intelligence with neuromorphic computing," *Nature,* vol. 575, no. 7784, pp. 607-617, 2019/11/01 2019, doi: 10.1038/s41586-019-1677-2.

[7] F. Zhou and Y. Chai, "Near-sensor and in-sensor computing," *Nature Electronics,* vol. 3, no. 11, pp. 664-671, 2020, doi: 10.1038/s41928-020-00501-9.

[8] S. S. Park and Y. S. Choi, "Spiking neural networks for physiological and speech signals: a review," (in eng), *Biomed Eng Lett,* vol. 14, no. 5, pp. 943-954, Sep 2024, doi: 10.1007/s13534-024-00404-0.

[9] Z. Yan, J. Zhou, and W.-F. Wong, "Energy efficient ECG classification with spiking neural network," *Biomedical Signal Processing and Control,* vol. 63, p. 102170, 2021/01/01/ 2021, doi: https://doi.org/10.1016/j.bspc.2020.102170.

[10] T. Chen, L. Wang, J. Li, S. Duan, and T. Huang, "Improving Spiking Neural Network With Frequency Adaptation for Image Classification," *IEEE Transactions on Cognitive and Developmental Systems,* vol. 16, no. 3, pp. 864-876, 2024, doi: 10.1109/TCDS.2023.3308347.

[11] G. Indiveri *et al.*, "Neuromorphic silicon neuron circuits," *Front Neurosci,* vol. 5, p. 73, 2011, doi: 10.3389/fnins.2011.00073.

[12] J. H. Baek *et al.*, "Artificial synaptic devices based on biomimetic electrochemistry: A review," *Materials Research Bulletin,* vol. 176, 2024, doi: 10.1016/j.materresbull.2024.112803.

[13] X. Ju and D. S. Ang, "Synapse and Tunable Leaky-Integrate Neuron Functions Enabled by Oxide Trapping Dynamics in a Single Logic Transistor," *IEEE Electron Device Letters,* vol. 43, no. 5, pp. 793-796, 2022, doi: 10.1109/led.2022.3162639.

[14] Q. Duan *et al.*, "Spiking neurons with spatiotemporal dynamics and gain modulation for monolithically integrated memristive neural networks," *Nat Commun,* vol. 11, no. 1, p. 3399, Jul 7 2020, doi: 10.1038/s41467-020-17215-3.

[15] Z. Wang *et al.*, "Memristors with diffusive dynamics as synaptic emulators for neuromorphic computing," *Nat Mater,* vol. 16, no. 1, pp. 101-108, Jan 2017, doi: 10.1038/nmat4756.

[16] S. Majumdar, "Back-End CMOS Compatible and Flexible Ferroelectric Memories for Neuromorphic Computing and Adaptive Sensing," *Advanced Intelligent Systems,* vol. 4, no. 4, 2021, doi: 10.1002/aisy.202100175.

[17] S. Dong, H. Liu, Y. Wang, J. Bian, and J. Su, "Ferroelectricity-Defects Synergistic Artificial Synapses for High Recognition Accuracy Neuromorphic Computing," *ACS Appl Mater Interfaces,* vol. 16, no. 15, pp. 19235-19246, Apr 17 2024, doi: 10.1021/acsami.4c01489.

[18] S. Majumdar, H. Tan, Q. H. Qin, and S. van Dijken, "Energy-Efficient Organic Ferroelectric Tunnel Junction Memristors for Neuromorphic Computing," *Advanced Electronic Materials,* vol. 5, no. 3, 2019, doi: 10.1002/aelm.201800795.

[19] S. Wang, D. W. Zhang, and P. Zhou, "Two-dimensional materials for synaptic electronics and neuromorphic systems," *Sci Bull (Beijing),* vol. 64, no. 15, pp. 1056-1066, Aug 15 2019, doi: 10.1016/j.scib.2019.01.016.

[20] W. Wang *et al.*, "Tailoring Classical Conditioning Behavior in TiO(2) Nanowires: ZnO QDs-Based Optoelectronic Memristors for Neuromorphic Hardware," *Nanomicro Lett,* vol. 16, no. 1, p. 133, Feb 27 2024, doi: 10.1007/s40820-024-01338-z.

[21] Y. R. Lee, T. Q. Trung, B. U. Hwang, and N. E. Lee, "A flexible artificial intrinsic-synaptic tactile sensory organ," *Nat Commun,* vol. 11, no. 1, p. 2753, Jun 2 2020, doi: 10.1038/s41467-020-16606-w.

[22] Y. Wei *et al.*, "Mechano-driven logic-in-memory with neuromorphic triboelectric charge-trapping transistor," *Nano Energy,* vol. 126, 2024, doi: 10.1016/j.nanoen.2024.109622.

[23] E. Chicca, F. Stefanini, C. Bartolozzi, and G. Indiveri, "Neuromorphic Electronic Circuits for Building Autonomous Cognitive Systems," *Proceedings of the IEEE,* vol. 102, no. 9, pp. 1367-1388, 2014, doi: 10.1109/JPROC.2014.2313954.

[24] B. V. Benjamin *et al.*, "Neurogrid: A Mixed-Analog-Digital Multichip System for Large-Scale Neural Simulations," *Proceedings of the IEEE,* vol. 102, no. 5, pp. 699-716, 2014, doi: 10.1109/JPROC.2014.2313565.

[25] H. Greatorex *et al.*, "A neuromorphic processor with on-chip learning for beyond-CMOS device integration," *Nature Communications,* vol. 16, no. 1, p. 6424, 2025/07/11 2025, doi: 10.1038/s41467-025-61576-6.

[26] M. Davies *et al.*, "Loihi: A Neuromorphic Manycore Processor with On-Chip Learning," *IEEE Micro,* vol. 38, no. 1, pp. 82-99, 2018, doi: 10.1109/MM.2018.112130359.

[27] O. Richter *et al.*, "DYNAP-SE2: a scalable multi-core dynamic neuromorphic asynchronous spiking neural network processor," *Neuromorphic Computing and Engineering,* vol. 4, no. 1, p. 014003, 2024/01/31 2024, doi: 10.1088/2634-4386/ad1cd7.

[28] S. Moradi, N. Qiao, F. Stefanini, and G. Indiveri, "A Scalable Multicore Architecture With Heterogeneous Memory Structures for Dynamic Neuromorphic Asynchronous Processors (DYNAPs)," *IEEE Transactions on Biomedical Circuits and Systems,* vol. 12, no. 1, pp. 106-122, 2018, doi: 10.1109/TBCAS.2017.2759700.

[29] J. Zhang *et al.*, "Spiking Neural Networks With Adaptive Membrane Time Constant for Event-Based Tracking," *IEEE Transactions on Image Processing,* vol. 34, pp. 1009-1021, 2025, doi: 10.1109/TIP.2025.3533213.

[30] Y. Zeng, E. Jeffs, T. Stewart, Y. Berdichevsky, and X. Guo, "Optimizing Recurrent Spiking Neural Networks with Small Time Constants for Temporal Tasks," presented at the Proceedings of the International Conference on Neuromorphic Systems 2022, Knoxville, TN, USA, 2022. [Online]. Available: https://doi.org/10.1145/3546790.3546796.

[31] A. Pazderka, "The role of membrane time constant in the training of spiking neural networks: Improving accuracy by per-neuron learning," Bachelor, Electrical Engineering, Mathematics and Computer Science, Delft University of Technology, The Netherlands, 2024. [Online]. Available: https://resolver.tudelft.nl/uuid:135b562c-d077-453c-a5d6-1a707da0659b

[32] T. Zhang, K. Yu, J. Zhang, and H. Wang, "DA-LIF: Dual Adaptive Leaky Integrate-and-Fire Model for Deep Spiking Neural Networks," in *ICASSP 2025 - 2025 IEEE International Conference on Acoustics, Speech and Signal Processing (ICASSP)*, 6-11 April 2025 2025, pp. 1-5, doi: 10.1109/ICASSP49660.2025.10888909.

[33] X. Li, S. Luo, and F. Xue, "Effects of synaptic integration on the dynamics and computational performance of spiking neural network," *Cognitive Neurodynamics,* vol. 14, no. 3, pp. 347-357, 2020/06/01 2020, doi: 10.1007/s11571-020-09572-y.

[34] P. V. A. Filippo Moro, Laura Kriener, Melika Payvand, "The Role of Temporal Hierarchy in Spiking Neural Networks," Institute of Neuroinformatics, UZH and ETH Zurich, Zurich, CH, 26 Jul 2024, 2024.

[35] B. Yin, F. Corradi, and S. M. Bohté, "Accurate and efficient time-domain classification with adaptive spiking recurrent neural networks," *Nature Machine Intelligence,* vol. 3, no. 10, pp. 905-913, 2021/10/01 2021, doi: 10.1038/s42256-021-00397-w.

[36] W. Fang, Z. Yu, Y. Chen, T. Masquelier, T. Huang, and Y. Tian, "Incorporating Learnable Membrane Time Constant to Enhance Learning of Spiking Neural Networks," *2021 IEEE/CVF International Conference on Computer Vision (ICCV),* pp. 2641-2651, 2020.

[37] G. D. Z. L. A. L. P. A. Beerel, "Spiking Neural Networks with Dynamic Time Steps for Vision Transformers," Ming Hsieh Dept. of Electrical and Computer Engineering, University of Southern California, Los Angeles, USA, 28 Nov 2023, 2023.

[38] A. B. N. Sowmya, "Design And Optimization Of Spiking Neural Networks For Energy-Efficient Neuromorphic Computing," *International Journal for Science and Advance Research In Technology,* vol. 11, no. 1, pp. 30-34, 07 Jan 2025 2025, Art no. IJSARTV11I1102616. [Online]. Available: https://ijsart.com/design-and-optimization-of-spiking-neural-networks-for-energy-efficient-neuromorphic-computing-102616.

[39] N. Rathi and K. Roy, "DIET-SNN: A Low-Latency Spiking Neural Network With Direct Input Encoding and Leakage and Threshold Optimization," *IEEE Trans Neural Netw Learn Syst,* vol. 34, no. 6, pp. 3174-3182, Jun 2023, doi: 10.1109/TNNLS.2021.3111897.

[40] R. A. B. Devine, H. N. Alshareef, and M. A. Quevedo-Lopez, "Slow trap charging and detrapping in the negative bias temperature instability in HfSiON dielectric based field effect transistors," *Journal of Applied Physics,* vol. 104, no. 12, 2008, doi: 10.1063/1.3039997.

[41] S. Majumdar, H. Tan, I. Pande, and S. van Dijken, "Crossover from synaptic to neuronal functionalities through carrier concentration control in Nb-doped SrTiO3-based organic ferroelectric tunnel junctions," *APL Materials,* vol. 7, no. 9, 2019, doi: 10.1063/1.5111291.

[42] S. Majumdar and I. Zeimpekis, "Back-End and Flexible Substrate Compatible Analog Ferroelectric Field-Effect Transistors for Accurate Online Training in Deep Neural Network Accelerators," *Advanced Intelligent Systems,* vol. 5, no. 11, p. 2300391, 2023, doi: https://doi.org/10.1002/aisy.202300391.

**Acknowledgement**

We acknowledge the support from the National Science and Technology Council of Taiwan (MOST-111-2222-E-006-005-MY2, NSTC-113-2221-E-006-138, NSTC-114-2221-E-006-205, NSTC-113-2410-H-006-118), the Higher Education Sprout Project, and the Ministry of Education to the Headquarters of University Advancement at National Cheng Kung University (NCKU). We also appreciate funding from the Taiwan Semiconductor Research Institute (JDP113-Y1-007 and JDP114-Y1-009), the 2024 New Faculty Equipment Usage Subsidy Project from the Core Facility Center at NCKU, and the AI SDGs Project from the Miin Wu School of Computing at NCKU. KB and SM acknowledge financial support from the Research Council of Finland through project AI4AI (Project number 350667) and Ferrari (no. 359047).

**Supplementary Materials**

# Task-Aware Tuning of Time Constants in Spiking Neural Networks for Multimodal Classification

Chiu-Chang Cheng[1], Kapil Bhardwaj[2], Ya-Ning Chang[1], Sayani Majumdar[1,2,*] and Chao-Hung Wang[1,3,*]

[1]Miin Wu School of Computing, National Cheng Kung University, 70101 Tainan City, Taiwan (R.O.C.)

[2]Faculty of Information Technology and Communication Sciences, Tampere University, 33720 Tampere, Finland

[3]Academy of Innovative Semiconductor and Sustainable Manufacturing, National Cheng Kung University, 70101 Tainan City, Taiwan (R.O.C.)

Corresponding authors: sayani.majumdar@tuni.fi; chwang@gs.ncku.edu.tw

In this work, a software-based approach was employed to evaluate the impact of the leaky time constant (LTC) on SNN performance. However, to investigate its hardware implications, the LTC must be correlated with practical hardware device characteristics. The conversion from the software time constant, $\tau_{discrete}$, to the hardware time constant, $\tau_{continuous}$, is demonstrated below:

**Conversion of software τ (discrete domain) to hardware τ (continuous domain)**

In the hardware model (continuous domain):

$$\tau_{continuous}\frac{dV}{dt} = -(V_t - V_{rest}) + I \qquad \text{(Eq. S1)}$$

Where $V$ denotes the membrane potential, $V_t$ is the membrane potential at time $t$, $V_{rest}$ is the resting potential, and $I$ is the input current.

In the software implementation (discrete domain):

$$V_{t+1} = V_t + \frac{V_{rest} - V_t}{\tau_{discrete}} + I \qquad \text{(Eq. S2)}$$

When t≈0, assume:

$$\frac{dV}{dt} \approx \frac{V_{t+1} - V_t}{\Delta t} \qquad \text{(Eq. S3)}$$

The correlation between the software implementation and hardware model becomes:

$$\tau_{continuous} \frac{V_{t+1} - V_t}{\Delta t} = -(V_t - V_{rest}) + I \qquad \text{(Eq. S4)}$$

This can be rearranged as:

$$V_{t+1} = V_t + \frac{(V_{rest} - V_t) * \Delta t}{\tau_{continuous}} + I * \frac{\Delta t}{\tau_{continuous}} \qquad \text{(Eq. S5)}$$

Therefore, the relationship between $\tau_{discrete}$ and $\tau_{continuous}$ is:

$$\tau_{discrete} = \frac{\tau_{continuous}}{\Delta t} \qquad \text{(Eq. S6)}$$

Example (ECG case with a sampling rate of 360 Hz):

In the software implementation, $\tau_{discrete}$ = 64 corresponds to a hardware time constant:

$$\tau_{continuous} = \frac{64}{360} \text{ seconds} \approx 0.1777 \text{ seconds}$$

Table 2 in the main text lists the detailed conversion of time constants from the discrete domain to the continuous domain.

**Network Architectures:**

The SNN architectures for the three different cases are presented below.

For static and dynamic image classification, a 2-layer fully connected SNN with 784 input neurons, 128 hidden neurons, and 10 output neurons was implemented, as shown in Figure S1. Direct input encoding was used, the time step was set to 10, and the dataset was the handwritten MNIST digits.

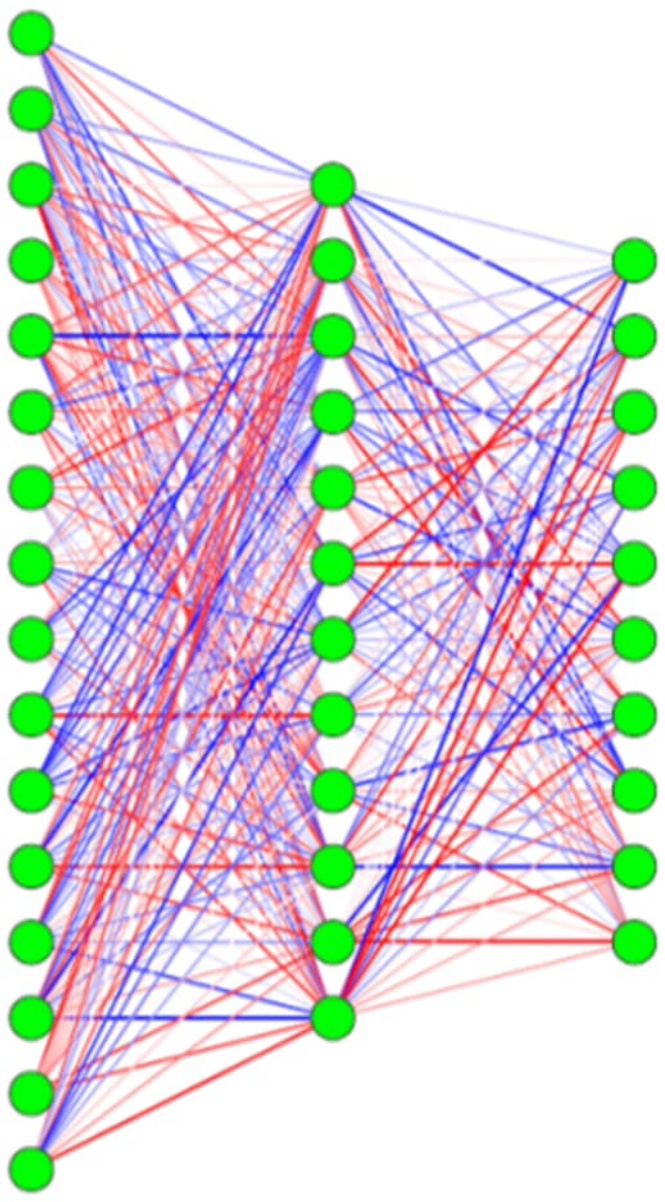

Figure S1. A 2-layer fully connected SNN is employed for static and dynamic image classification.

For time-series data classification, a 4-layer fully connected SNN was established, consisting of 1 input neuron, 784, 256, and 64 hidden neurons, and 4 output neurons, as shown in Figure S2. Direct input encoding was used, and the ECG dataset was employed.

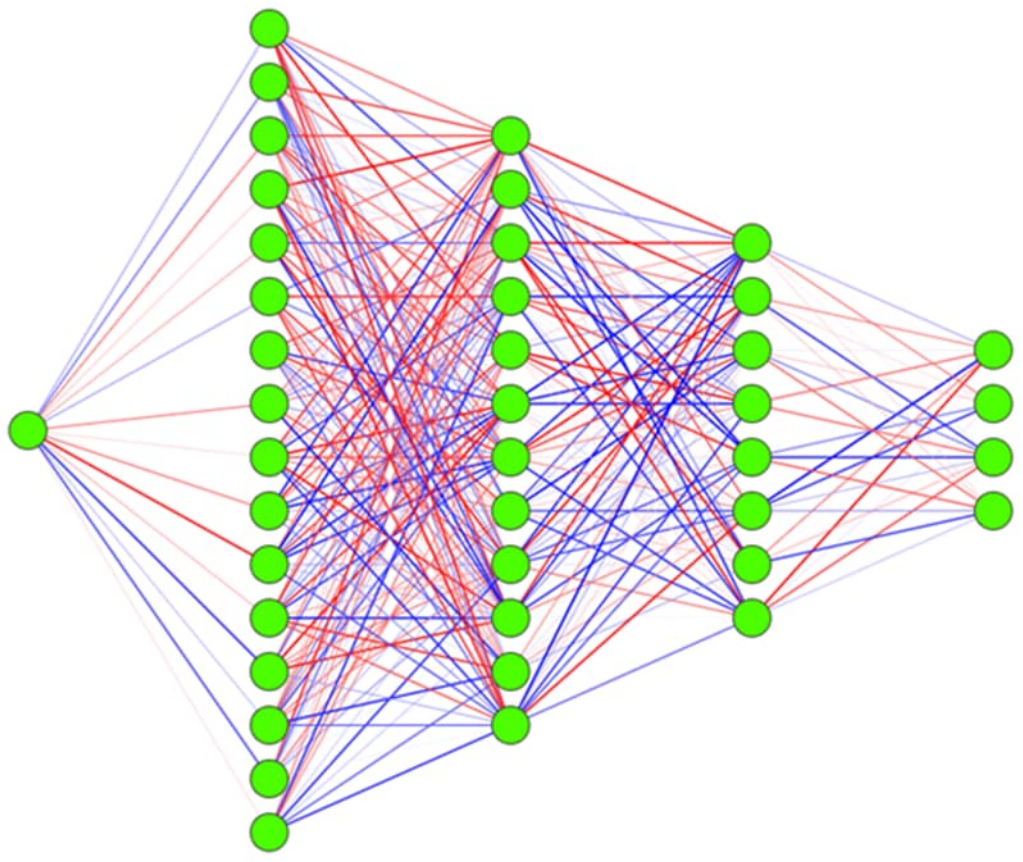

**Figure S2.** A 4-layer fully connected SNN was implemented for time-series data classification.